\pdfoutput=1

\documentclass[11pt,a4paper]{article}

\AtBeginDocument{%
  \providecommand\BibTeX{{%
    \normalfont B\kern-0.5em{\scshape i\kern-0.25em b}\kern-0.8em\TeX}}}

\usepackage{acl}

\usepackage{times}
\usepackage{latexsym}

\usepackage[T1]{fontenc}

\usepackage[utf8]{inputenc}

\usepackage{microtype}
\usepackage{hyperref}
\usepackage{graphicx}
\usepackage{amsmath}
\usepackage{adjustbox}
\usepackage{booktabs}
\usepackage{subfig}
\usepackage{tabularx}
\usepackage{enumitem}
\usepackage[depth=subsection]{bookmark}
\setlist[enumerate]{topsep=0pt}
\title{Examining Bias in Opinion Summarisation Through the Perspective of Opinion Diversity}
\author{
\begin{tabular}{c c c c}
    Nannan Huang & Lin Tian &
    Haytham Fayek & Xiuzhen Zhang \\
\end{tabular}
\\[5pt]
RMIT University, Australia
\\[5pt]
\texttt{\{amber.huang, lin.tian2\}@student.rmit.edu.au} \\
\texttt{haytham.fayek@ieee.org} \\
\texttt{xiuzhen.zhang@rmit.edu.au} \\
}
\begin{document}
\maketitle
\begin{abstract}
Opinion summarisation is a task that aims to condense the information presented in the source documents while retaining the core message and opinions.
A summary that only represents the majority opinions will leave the minority opinions unrepresented in the summary. 
In this paper, we use the stance towards a certain target as an opinion. 
We study bias in opinion summarisation from the perspective of opinion diversity, 
which measures whether the model generated summary can cover a diverse set of opinions.
In addition, we examine opinion similarity, a measure of how closely related two opinions are in terms of their stance on a given topic, and its relationship with opinion diversity.
Through the lens of stances towards a topic, we examine opinion diversity and similarity using three debatable topics under COVID-19. Experimental results on these topics revealed that a higher degree of similarity of opinions did not indicate good diversity or fairly cover the various opinions originally presented in the source documents. 
We found that BART \cite{lewis2020bart} and ChatGPT can better capture diverse opinions presented in the source documents.
\end{abstract}

\section{Introduction}
The aim of opinion summarisation is to reduce the amount of information in the source text while maintaining the core message and opinions expressed therein. It can be in the form of a summary of a product review \cite{alam2016joint, bravzinskas2020unsupervised, chu2019meansum}, online discourse using platforms like Twitter, Reddit \cite{fabbri-etal-2021-convosumm, bilal-etal-2022-template} or other types of text with opinions. 
There are two major types of models, extractive \cite{meng2012entity, ku2006opinion, erkan2004lexrank, liu2019fine} and abstractive \cite{lewis2020bart, raffel2020exploring, zhang2020pegasus}. The extractive models extract the key information by selecting the most salient sentences in the source documents. Whereas the abstractive summarisation models generate new phrases that reflect the key information in the source documents. Applications for this activity include tracking consumer opinions, assessing political speeches, or internet conversations, among many others.

A summarisation model's output will reflect any biases present in the data used for training the model. Moreover, summarisation models are used in many applications where the fairness of the outputs is critical. For instance, opinion summarisation models can be used to summarise product reviews and social media posts. If these models produce biased summaries, they have the risk of being misused as a tool to influence and manipulate people's opinions. Therefore, it is essential to look into the fairness of the models.


Earlier studies of bias in opinion summarisation have mainly evaluated biases in summarisation by comparing whether the information is selected from different social groups equally, using attributes including gender, dialect, and other societal properties \cite{dash2019summarizing, blodgett-etal-2016-demographic, keswani2021dialect, olabisi-etal-2022-analyzing}. Such representation is fair from the perspective of information input, leaving the other side of information consumption under studied. It is equally important to look at fairness from end-users' perspective. \citet{9378095} pointed out that fair representation from the information input perspective does not always imply fair representation from end-users' perspective. From the standpoint of end-users, it is more important for summaries to cover the various opinions presented in the source documents so that the various opinions in the source documents can be heard equally \cite{9378095}. In this study, we examine bias in opinion summarisation from end-users' perspective by comparing the output of different modern summarisation models.

\citet{blodgett-etal-2020-language} noted that many research work on bias in natural language processing lacks a precise description and definition of \textit{bias} in their study, making it challenging for readers to understand.
Our working definitions of the key terms are as follows.
According to the \textit{stance triangle theory} \cite{du2007stance}, a stance is composed of three components: the object, the subject, and the attitude, which shape sociocultural value systems and opinion expression. 
While studies such as \citet{misra-etal-2016-measuring} and \citet{reimers-etal-2019-classification} utilise arguments with similarity and clustering techniques to capture similar opinions, our study takes a different approach based on the stance triangle theory. In our study, an \textit{opinion} is a personal stance about a particular target, and a stance is a position or attitude that an individual takes about a given target.
Our definition of \textit{bias in opinion summarisation} is when the summary focuses more on certain opinions than on the diversity of opinions presented in the source documents. Note that it is generally agreed that diverse opinions should be taken into account regardless of their quantitative value in order to include more diverse opinions when using sentiment information \cite{angelidis2018summarizing, siledar2023aspect}. Hence, our focus is on measuring the diversity of opinions rather than quantity. This is measured through \textit{opinion diversity} which assesses opinion equivalence relative to the source documents. It measures the opinions in the source documents that the generated summary contains.
In addition, we compare opinion similarity between the source and generated documents and further examine the relationship between opinion diversity and opinion similarity. \textit{Opinion similarity} is a measure of how closely related two opinions are in terms of their stance on a given topic. We use BERT \cite{devlin2019bert} to compare the semantic closeness in the embedding space. 
We aim to understand whether models perform well in capturing overall opinions that are less biased by covering diverse opinions. We examine opinions on three COVID-19 topics using stances on these topics. 

In our study, we aim to answer the following questions:
\begin{enumerate}[itemsep=0pt, parsep=0pt]
\item How well can summarisation models present various opinions in a document from the perspective of stance towards a topic?
\item Does a greater degree of opinion similarity with the source documents suggest a lack of bias in summarisation models?
\end{enumerate}

\section{Related Work}
\subsection{Opinion summarisation}

Opinion summarisation has received significant attention in recent years, with extractive models such as Hybrid-TFIDF \cite{inouye2011comparing} a frequency-based summarisation method, having great performance for summarising social media data like tweets. Recent studies have introduced the concept of key point analysis \cite{bar2020arguments, bar2020quantitative, bar2021every}, which uses extractive models to identify key arguments from the source documents and match them to the main opinions' associated key points. Abstractive opinion summarisation models, such as Copycat \cite{bravzinskas2020unsupervised} and MeanSum \cite{chu2019meansum}, are designed to address the problem of summarising product or business reviews. MeanSum \cite{chu2019meansum} is based on LSTM, while Copycat \cite{bravzinskas2020unsupervised} uses a variational autoencoder (VAE) to generate latent vectors of given reviews.



\subsection{Biases in opinion summarisation}
Existing studies of bias in opinion summarisation have focused on the perspective of using sensitive attributes of social media users and categorising them under different social groups \cite{dash2019summarizing}. These attributes include social identities like gender, race and political leaning information. Other studies focused on the perspective of dialect used in text and whether the generated summaries cover such dialects \cite{blodgett-etal-2016-demographic, keswani2021dialect, olabisi-etal-2022-analyzing}.



\begin{figure*}[tbh]
    \centering
    \includegraphics[width=12cm]{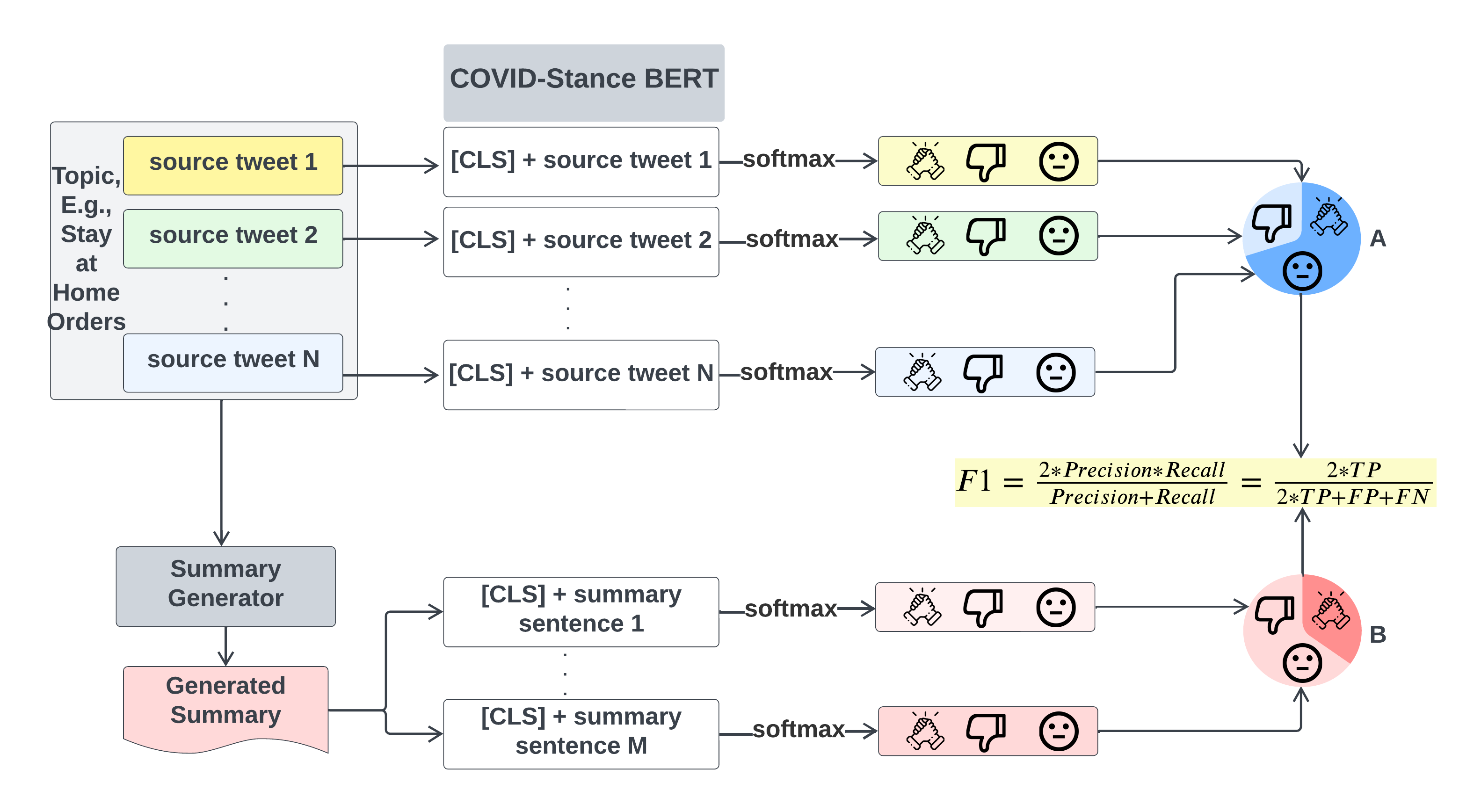}
    \caption{The process of calculating opinion diversity: COVID-Stance-BERT is applied to each source tweet and aggregated into a single set of opinions. The same process is applied to the generated summaries at the sentence level. The F1 score is applied among the two sets of opinions, where the true positive (TP) measures opinions captured by the model generated summary that is also presented in the source documents; false positive (FP) measures opinions captured by the model generated summary that is not presented in the source documents; false negative (FN) measures opinions not captured by the model generated summary that is presented in the source documents.}
    \label{fig:various_opinion_comparison}
\end{figure*}



One limitation of these approaches is that they require obtaining sensitive hidden attributes from the text producers. Due to other factors including privacy, this is not always practicable. Another limitation, as pointed out in \citet{9378095} is the assumption of fairness in using information from different social groups equally 
and this is from the input information perspective. Fairness should also be evaluated from the perspective of end-users. \citet{9378095} discovered in their study that, from the perspective of end-users, it is more important to present diverse information in the source documents. We argue that considering fairness from the perspective of end-users is equally important since they are the users of the final product and make informed decisions based on the generated summaries.

Therefore, in our work, we focus on the fair representation of information generated by summarisation models from end-users' perspective. 
We study bias in opinion summarisation by examining whether the summary focuses more on certain opinions than on the diverse opinions presented in the source documents. 

\section{Methodology}

As previously stated, we capture opinion in our study by utilising stance and its target. When the generated summaries reflect opinions that diverge from those of the source texts, the summary is considered biased. This could take the form of concentrating on a narrow range of opinions or going beyond what was expressed in the source documents. It should be noted that the biased information in the source text is not the main focus of our study; rather, we are interested in how opinions are presented and whether the opinions in the generated summaries match those in the source documents.

We formulate our problems in three steps. We first use a pre-trained language model to capture opinions from the input sequences. 
The opinion diversity is then calculated using the F1 score between the set of stances in the source tweets and the generated summary for each cluster (discussed below) under each topic. 
Finally, we compare the cosine similarity between the source tweet cluster and the summary at the sentence level to measure opinion similarity using model representation.

Let $\mathcal{C} = \{c_0, c_1, c_2,...,c_l\} $ be a set of clusters. These clusters are derived from the three main topics (``CDC'', ``Stay at Home Orders'' and ``Wearing a Face Mask'').
For each $c$, we have a set of source tweets $\mathcal{T}$, defined as $\mathcal{T} = \{t_0, t_1, t_2,...,t_n\}$.
For each $c$, we have a set of generated summaries $\mathcal{S} = \{\mathbf{s}_0,\mathbf{s}_1,\mathbf{s}_2,...,\mathbf{s}_l\}$, where each summary $\mathbf{s}$ consists of a lists of sentences, defined as $\mathbf{s} = \{ e_0, e_1, e_2,...,e_q\}$, where $e$ refers to the textual content of each input sentence.

\subsection{Capturing Opinions}
We train COVID stance classification models related to several COVID-19 topics or targets using a publicly available dataset from \citet{glandt2021stance} on COVID-19 related stance detection. The dataset consists of four different COVID-19 related topics and targets. See Appendix~\ref{sec:stance_dataset} for further detail on the summary of the data distribution.
Similar to \citet{glandt2021stance}, using the further pre-trained BERT model (bert-large-uncased)\footnote{\url{https://huggingface.co/bert-large-uncased}} with the COVID-19 tweet corpus~\cite{muller2020covid}, we fine-tune the model with stance labeled data from~\citet{glandt2021stance}. Thus, our COVID-Stance-BERT is a further pre-trained BERT fine-tuned with standard cross entropy loss to do a three-class classification of stance labels (support, against, and neutral). 
Each tweet $t_i$ is associated with a ground-truth label $d \in \mathcal{D}$, where $\mathcal{D}$ represents the label set (support, against or neutral; 3 classes).
\begin{align}
    v_i &= \text{BERT}([\text{CLS}] \oplus x_i )\label{eqn:bert} \\ 
    \hat{d_i} &= \text{softmax} (W v_i + b)  \label{eqn:softmax}
\end{align}
We call these models COVID-Stance-BERT in the remainder of the paper. The average accuracy and macro F1 scores across three targets are 0.8208 and 0.8026 respectively. Similar levels of accuracy and F1 scores were obtained across these topics compared to \citet{glandt2021stance}. The detailed result is reported in Appendix~\ref{sec:covid_stance_bert_performance}.


\subsection{Opinion Diversity}
\label{opinion_coverage}
The overall process is presented in Fig~\ref{fig:various_opinion_comparison}. 
In order to determine whether different opinions stated in the source documents can be captured using summarisation models, we apply COVID-Stance-BERT to the source documents to produce a collection of opinions represented in the input tweets and the generated summaries.
The majority of the source tweets have only a single sentence. We therefore treat each sentence in the generated summaries as a tweet and apply the same COVID-Stance-BERT to retrieve its opinions. We apply the in-domain stance detection model on both the source documents and the summarised sentences across the tweet clusters under different topics. 
That is, a stance detection model fine-tuned to the target of "Stay at Home Orders" is applied to the collection of tweets and the generated summaries towards the "Stay at Home Orders" topic. 

Once the prediction is done on both the source documents and the generated summaries, we compare the sets of opinions and examine if the generated summaries cover the various opinions presented in the source document.
We compute the F1 score by comparing the two sets of stances under each discussed topic across all clusters and use them to represent how well the summarisation model captures the stances in relation to the input documents.


The detailed calculation can be found as follows:
we apply COVID-Stance-BERT stance prediction to calculate the opinion diversity on both the set of source tweets and all generated summaries.
For each generated summary ${s_j} \in \mathcal{S}$, we adopt sentence splitting function\footnote{\url{https://www.nltk.org/api/nltk.tokenize.html}} as $s_j = \{ss_0,ss_1,...,ss_m\}$ 
For each sentence $e$ in generated summary $s_j$, we take the associated stance label $d_e$,
formally,
\begin{equation}
\begin{aligned}
    d_{t} = \text{BERT}(\text{emb}([CLS], t_p)), \\ 
    d_{ss} = \text{BERT}(\text{emb}([CLS], ss_q)),
\end{aligned}
\end{equation}
where $t_p$ represents the text of the source tweet and $s_q$ is the summary, $\text{emb}()$ the embedding function and 
$d_{t}$ and $d_{ss}$ are the stance predictions based on the $[CLS]$ token produced by our COVID-Stance-BERT.



Once the above is completed, we get the set of non-repeated opinion(s) in both the source documents and the generated summaries.
We use the F1 score, which measures the harmonic mean of opinion precision and opinion recall, to evaluate the performance of opinion diversity, where the opinion precision measures the proportion of important opinions in the generated summary. The opinion recall measures the degree of salient opinion in the source documents that the generated summary contains.
A higher F1 score indicates the model generated summary can better cover the various opinions presented in the source documents. The true positive (TP), false positive (FP), and false negative (FN) are measured as follows:
\begin{itemize}
    \item TP = opinions captured by the model generated summary that is also presented in the source documents.
    \item FP = opinions captured by the model generated summary that is not presented in the source documents.
    \item FN = opinions not captured by the model generated summary that is presented in the source documents.
\end{itemize}
More detail on illustrations of various scenarios of opinion precision and recall and their associated F1 scores can be found in Appendix~\ref{sec:illustration_opinion_diversity}.
We report the average across all clusters under each topic as the overall opinion diversity for each model.

\subsection{Opinion Similarity}
\begin{figure*}[tbh]
    \centering
    \includegraphics[width=12cm]{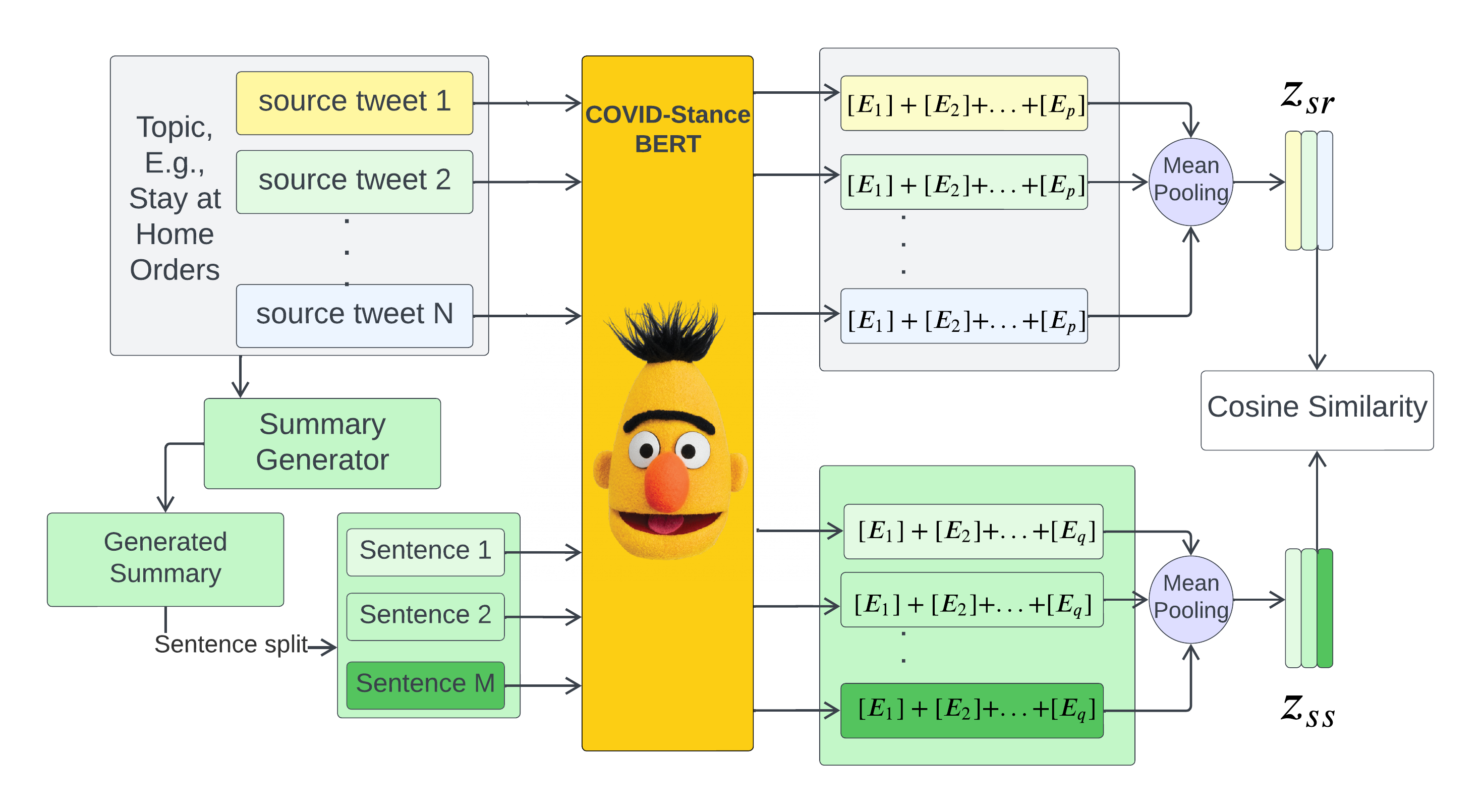}
    \caption{The process of calculating opinion similarity.
    For the source tweets, a single representation is obtained using mean-pooling approach based on the mean representations of all input tokens.
    For the generated summaries, the same process is applied at the sentence level. The cosine similarity is then applied to these representations to measure opinion similarity.
    }
    \label{fig:cls_cosine_similarity}
\end{figure*}
The overall process of evaluating opinion similarity is visualised in Fig~\ref{fig:cls_cosine_similarity}. 
To aggregate all the source tweets in a cluster to get a source representation ($z_{sr}$), we adopt the mean-pooling approach based on the mean representations of all input tokens:
\begin{equation}
    z_{sr} = \frac{1}{|n|} \sum_{i \in n} h_i^c,
    \label{eqn:source_rep}
\end{equation} where $h_i^c$ is the mean stance representation derived from source tweet $t_i$.

To extract the representation for each generated summary ($z_{ss}$), we applied the mean-pooling approach over the average of all token representations from the last layer of our language model encoder as shown in Eq.~\ref{eqn:summary_rep}:
\begin{equation}
    z_{ss} = \frac{1}{|m|} \sum_{i \in m} v_i^c,
    \label{eqn:summary_rep}
\end{equation} where $v_i^c$ is the mean stance representation based on each token representation from a given summary $s_i$.

For each generated summary, we compute the cosine similarity to the average source representation in the cluster, with the following equation:
\begin{equation} 
\cos (z_{sr},z_{ss})= {z_{sr} z_{ss} \over \|z_{sr}\| \|z_{ss}\|}.
\end{equation}
 A higher cosine similarity score between the source and the summary indicates a model is better at capturing the overall opinions and semantic information presented in the source documents.


In summary, we investigate bias in opinion summarisation models from the standpoint of opinion diversity. The idea is to look at whether summarisation models can capture the various opinions presented in the source documents. In addition, from the perspective of opinion similarity, we evaluate how closely related the generated summary and the source documents are in terms of their stance and semantic information on a given topic. Combining the results of opinion diversity and opinion similarity, we aim to understand whether summaries that express opinions that are overall similar to the source documents also indicate they are less biased.

\section{Experiments and Results}

\subsection{Data}

In this study, we use the COVID-19 tweet clusters dataset provided by \citet{bilal-etal-2022-template}. 
The dataset contained tweet clusters that are coherently opinionated, coherently non-opinionated, and incoherent subsets. 
Each topic contains a different number of clusters. 
A cluster of tweets contains a collection of tweets obtained during a particular timestamp on a specific topic. 
In \citet{bilal-etal-2022-template} each cluster was used for generating the final summary. Hence, in our work, we focus on the cluster level when generating summaries for each topic.
For example, under the topic "Wearing a Face Mask" there can be multiple clusters obtained at different times; each cluster of tweets could be discussing whether wearing a face mask is a good idea, obtained at a different time of the year.

In our experiment, we are only considering tweet clusters that are coherently opinionated, with a similar discussion of targets as the ones mentioned in \citet{glandt2021stance}.
This is to ensure that all tweets contain opinions and, at the same time, to utilise the stance detection model in an in-domain setting to evaluate the opinions expressed in these clusters. We obtained coherently opinionated clusters including "Stay at Home Orders", "Wearing a Face Mask" and a highly related topic, "CDC/Centres for Disease Control and Prevention", where the clusters of tweets centred around the aforementioned topics are mainly focusing on the discussion and the expression of opinions towards them during the COVID-19 pandemic.
The in-domain stance detection models were then applied to the clusters apart from "CDC/Centres for Disease Control and Prevention" where we applied the stance detection model that was trained on the "Fauci" topic since they are both public figures.

The overall data distribution after our selection can be found in Table~\ref{tweet_cluster_distribution}. There are 78, 48, and 52 clusters of tweets; and on average, 21.77, 20.54, and 22.42 tweets under each cluster for the "CDC", "Stay at Home Orders", and "Wearing a Face Mask" topics, respectively. 

\begin{table}
  \caption{The statistic of the number of tweet clusters and the average number of tweets in each cluster under different topics.}
  \label{tweet_cluster_distribution}
{\fontsize{10}{12}\selectfont
    \begin{tabular}{p{3cm}p{1.5cm}p{1.5cm}}
    \toprule 
    Topic & No. Clusters & Ave No. Tweets \\ 
    \midrule
        CDC & 78 & 21.77\\
        Stay at Home Orders & 48 & 20.54\\
        Wearing a Face Mask & 52 & 22.42\\
    \bottomrule
    \end{tabular}}

\end{table}

\subsection{Baseline Models}
\label{baseline_models}
Several summarisation models are used to generate summaries in our experiments, including extractive summarisation models TextRank \cite{mihalcea2004textrank}, LexRank \cite{erkan2004lexrank} and Hybrid-TFIDF \cite{inouye2011comparing}; and abstractive summarisation models BART \cite{lewis2020bart}, Pegasus \cite{zhang2020pegasus} and T5 \cite{raffel2020exploring}, a summarisation model for review Copycat \cite{bravzinskas2020unsupervised} and a recently released Large Language Model(LLM) - ChatGPT\footnote{\url{https://chat.openai.com/}}.
Following \citet{bilal-etal-2022-template} we limit the abstract summarisation models word limit to the generated summary within [90\%, 110\%] of the gold standard length; 
 and the average token length of the gold standard length for ChatGPT. 
For extractive models only allows selecting sentences we limit to the average number of sentences of the gold standard length.
We apply all the models mentioned in a zero-shot setting.
More in-depth discussion on each of the models is below:
\begin{itemize}
    \item \textbf{BART} \cite{lewis2020bart} is an encoder-decoder model with a bidirectional encoder and a left-to-right decoder.
    Pretrained using a novel in-filling technique by replacing a span of text with a single mask token. Making it useful for language generation tasks
    We use the BART large model, pre-trained on CNN/Daily Mail
    \footnote{\url{https://huggingface.co/facebook/bart-large-cnn}}.
    
    \item \textbf{Pegasus} \cite{zhang2020pegasus} is a model that employs the Transformer encoder and decoder, self-supervised learning and pretraining on predicting the removed sentences, and tokens similar to the masked language model. 
    We use the Pegasus model pre-trained on CNN/Daily Mail
    \footnote{\url{https://huggingface.co/google/pegasus-cnn_dailymail}}.

    \item \textbf{T5} \cite{raffel2020exploring} is an encoder-decoder model pretrained on a multi-task setting using both supervised and unsupervised settings where the tasks are converted into a set of input-output text pairs. This allows it to understand a large variety of relationships between texts.
    We use the T5 base model pre-trained on CNN/Daily Mail 
    \footnote{\url{https://huggingface.co/flax-community/t5-base-cnn-dm}}.
\begin{table*}[tbh]
  \caption{Results of opinion diversity (Opi Div) and opinion similarity (Opi Sim) for various models under different discussed topics. 
  The best results are bolded, and the ranking of the models is provided inside the brackets.}
  \label{result_table}
    \centering
    {\fontsize{10}{12}\selectfont
    \begin{tabular}{c c c c c c c}
    \toprule
     Events & 
    \multicolumn {2}{c}{CDC} &
    \multicolumn {2}{c}{Stay at Home Orders} &
    \multicolumn {2}{c}{Wearing a Face Mask} \\
    \midrule
    Models & Opi Div & Opi Sim & Opi Div & Opi Sim & Opi Div & Opi Sim\\
    \midrule
        BART	&	\textbf{0.7449	(1)}	&	0.8503	(4)	&	0.7681	(2)	&	0.8373	(7)	&	\textbf{0.8147	(1)}	&	0.8412	(6)	\\
        Pegasus	&	0.5265	(7)	&	0.8745	(3)	&	0.7576	(3)	&	0.8775	(3)	&	0.3692	(5)	&	0.8768	(3)	\\
        T5	&	0.6346	(3)	&	0.8451	(5)	&	0.7417	(5)	&	0.8407	(6)	&	0.4692	(4)	&	0.8327	(7)	\\
        ChatGPT	&	0.7282	(2)	&	0.8818	(2)	&	\textbf{0.8014	(1)}	&	0.8515	(5)	&	0.6006	(3)	&	0.8498	(5)	\\
        Copycat	&	0.5265	(7)	&	0.6725	(8)	&	0.7014	(8)	&	0.7288	(8)	&	0.6737	(2)	&	0.7177	(8)	\\
        TextRank	&	0.5338	(6)	&	0.8370	(6)	&	0.7417	(5)	&	0.8519	(4)	&	0.2615	(8)	&	0.8828	(2)	\\
        LexRank	&	0.5530	(5)	&	0.8208	(7)	&	0.7569	(4)	&	0.8817	(2)	&	0.3590	(7)	&	0.8607	(4)	\\
        Hybrid TFIDF	&	0.5697	(4)	&	\textbf{0.8914	(1)}	&	0.7063	(7)	&	\textbf{0.8923	(1)}	&	0.3667	(6)	&	\textbf{0.8965	(1)}	\\
    \bottomrule
    \end{tabular}}
\end{table*}
    \item \textbf{ChatGPT} 
    OpenAI's ChatGPT, a recently released Large Language Model (LLM), was developed by employing reinforcement learning from human feedback (RLHF) \cite{christiano2017deep} to train a GPT-3.5 series model.
    We employ OpenAI's ChatGPT API (gpt-3.5-turbo-0301) for our experiments. 
    We adjusted the maximum tokens to the average gold standard token length. 
    We prompt ChatGPT as below "Summarise the following tweets: You should write it in tweet style. You should use no more than 4 sentences. Tweets: [Source tweets]".

    \item \textbf{Copycat} \cite{bravzinskas2020unsupervised} is a Variational Autoencoder model overcomes issues with limited training data in the context of review summarisation using self-supervision and latent representation that represents the general opinions expressed in the source reviews. We use the model provided by the authors
    \footnote{\url{https://github.com/abrazinskas/Copycat-abstractive-opinion-summarizer}.}.
    
    \item \textbf{TextRank} \cite{mihalcea2004textrank} is a graph-based model that extracts the most important sentence from the input document based on the weight determined by the edges connected to the words or phrases.

    \item \textbf{LexRank} \cite{erkan2004lexrank} is a graph-based model represents each sentence in the document as a node. Edges between vertices are calculated using cosine similarity, and the importance of a sentence is determined by the number of connected edges. The model extracts the most important sentence from the document based on the connectivity matrix.

    \item \textbf{Hybrid-TFIDF} \cite{inouye2011comparing} is a graph-based model similar to TextRank \cite{mihalcea2004textrank} and LexRank \cite{erkan2004lexrank}. Each word in the sentence is represented using the TF-IDF score, and similarity between sentences is computed to build edges between sentence vertices. Similar to other graph-based models, sentences with the most connected edges are deemed most important.
\end{itemize}

\begin{figure*}[tbh]
    \centering
    \includegraphics[width=14cm]{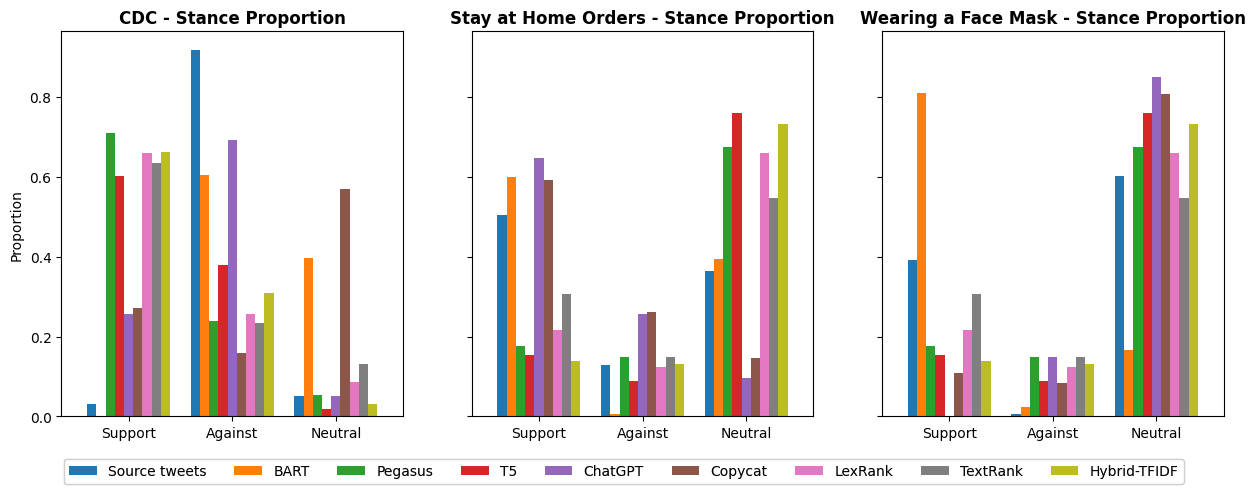}
    \caption{Stance distribution across various discussion topics for source documents and model generated summaries.}
    \label{stance_distribution}
\end{figure*}

\subsection{Opinion Diversity by Models}
\label{opinion_diversity}
To answer the first question, we first gather opinions from the source input tweets by running COVID-Stance-BERT on each tweet in each cluster under different topics. Similarly, the same model is used to collect opinions captured by different summarisation models as discussed in Section~\ref{opinion_coverage}. Once the sets of opinions from the source document and the generated summaries are obtained, we compare whether the opinions presented in the source document are also the opinions captured by the models. This is to answer the question of whether models can fairly preserve the various opinions being presented in the source documents.

We compare the three different discussed topics listed in Table~\ref{tweet_cluster_distribution} using the clusters within each topic. Summaries were first obtained using the models mentioned in Section~\ref{baseline_models} for each cluster. Then, using the opinion sets obtained for both the source documents and the generated summaries, we compare whether the summaries have high opinion diversity by comparing the set of stance(s) captured by the summarisation models with the set of stance(s) presented in the source documents.
See Appendix~\ref{sec:example_opinion_diversity_calculation} for details of the stance detection example and an example calculation of opinion diversity.

The results of the opinion diversity of different models under different discussion topics are listed in Table \ref{result_table}.
We are reporting the F1 score for opinion diversity. The results of opinion precision and recall can be found in Appendix~\ref{sec:additional_results}.
BART \cite{lewis2020bart} has the best opinion diversity for both the "CDC" and "Wearing a Face Mask" topics, while it has a competitive result for the "Stay at Home Orders".
ChatGPT has the best opinion diversity for "Stay at Home Orders" and is relatively competitive for the other topics.

Overall speaking, BART has the best performance in capturing diverse opinions and followed by ChatGPT. 
It is worth noticing, however, that previous studies found that ChatGPT tends to generate lengthy summaries when using the default parameters \cite{wang2023cross, yang2023exploring} and it is important to provide guidance in the prompt. The fact that we limit its maximum tokens and ask it to generate a fixed number of sentences may hurt its performance. For example, in the generated summary presented in Appendix~\ref{sec:additional_results}, ChatGPT generated an unfinished sentence.


\subsection{Opinion Similarity by Models}
\label{opinion_similarity}

The second question we aimed to answer is whether a model that generates summaries with high overall opinion similarity to the source documents would also be less biased by covering various opinions. To answer this question, we used the average input token representation obtained using the COVID-Stance-BERT model, for both the source documents and also the generated summaries. 

For the source documents, the average input token representation for each tweet was obtained using COVID-Stance-BERT, followed by applying mean-pooling to aggregate them into a single representation. For the summary, a similar approach was applied to the generated summary at the sentence level. We obtained average input token representations for each sentence in the generated summary, and then a mean pooling was applied over the sentence representations to obtain a single representation for the generated summary. Once the representations are obtained for both the source documents and the model generated summaries. We compare the stance similarity between source documents against different model generated summaries using the cosine similarity between the source representation and the summary representation.

Table \ref{result_table} shows the cosine similarity for measuring how similar the overall stance presented in the generated summaries is to the source documents. From the result, we can see that the model with the overall highest similarity of opinions across different topics is Hybrid-TFIDF \cite{inouye2011comparing}. We suspect this can be due to the superior performance of Hybrid-TFIDF \cite{inouye2011comparing} in summarising microblogging text. This is most likely because Twitter posts are not like typical documents and are unstructured, disconnected, and brief \cite{inouye2011comparing}. As a result, when compared to the source documents, it maintains a similar overall stance. 

\subsection{Opinion Distribution}
\label{opinion_distribution}


We computed the stance distribution across the discussion topics using the proportion of stances in the source documents and different summarisation models. The outcome is depicted in Fig.~\ref{stance_distribution}. 
From the distribution result, we can see that BART \cite{lewis2020bart} can better preserve the stance distribution for both the "CDC" and "Wearing a Face Mask" topics, whereas ChatGPT can better preserve the stance distribution for the "Stay at Home Orders" topic. While most models could pick up various stances and present a similar stance distribution compared to the source documents for the "Stay at Home Orders" topic. We suspect this is due to the fact that this topic has no obvious minority opinion, making it less challenging for models to cover diverse opinions. The above aligns with what we found in Section~\ref{opinion_diversity} where BART \cite{lewis2020bart} and ChatGPT outperformed the other models by covering more diverse opinions. 
We believe one of the possible reasons could be due to BART's impressive multi-document summarisation capabilities \cite{chen2020multi, johner2021error} and ChatGPT's good performance in multiple downstream NLP tasks. 

In conclusion, based on the results in Section \ref{opinion_diversity} and \ref{opinion_similarity} we found that when the generated summary has a higher degree of similarity in terms of overall opinion, that does not indicate it is fair in terms of covering more diverse opinions. In combination with the result from Section \ref{opinion_distribution} we observed that the model's ability to capture various opinions on different topics is case-dependent. We found that when no obvious minority stance is presented in the source documents, most models could capture various opinions in the source document.

\section{Conclusion}
In this work, we proposed a new way to examine bias in opinion sumamrisation from the perspective of presenting various opinions in the summary. We investigated various summarisation models for the COVID-19 event under three topics, using stance towards a target as a representation of opinions. In addition, we also examined overall stance similarity using model representation.
We found that BART \cite{lewis2020bart} and ChatGPT are better at capturing diverse opinions when generating a summary. Whereas Hybrid-TFIDF has the highest similarity across the three discussed topics for overall opinion similarity.
Based on the result, we found that higher opinion similarity does not indicate that the model presents diverse opinions. While both attributes are important for evaluating bias in opinion summarisation
we suggest future studies look into introducing metrics that can evaluate summaries from both perspectives.

\bibliographystyle{acl_natbib}
\bibliography{anthology, main}

\appendix
\section{Appendix}

\subsection{Dataset}
\label{sec:stance_dataset}
The dataset provided by \citet{glandt2021stance} consists of four different stance targets related to COVID-19. The detail of the data distribution can be found in Table~\ref{covid_stance_data_distribution}. 

\begin{table}[tbh]
  \caption{COVID-related stance detection data distribution in training, validation and test subsets per target provided in \citet{glandt2021stance}}
  \label{covid_stance_data_distribution}
    \begin{tabular}{c c c c}
    \toprule
    Target & Train & Val & Test\\
    \midrule
    Anthony S. Fauci, M.D. & 1464 & 200 & 200\\
    Keeping Schools Closed & 790 & 200 & 200\\
    Stay at Home Orders & 972 & 200 & 200\\
    Wearing a Face Mask & 1307 & 200 & 200\\
    \bottomrule
\end{tabular}
\end{table}

\subsection{COVID-Stance-BERT Performance}
\label{sec:covid_stance_bert_performance}
\begin{table}[tbh]
  \caption{Performance of the COVID-Stance-BERT models for stance detection on the targets in the dataset provided by \citet{glandt2021stance}. The performance is reported in terms of accuracy (Acc) and macro F1 score (F1).}
  \label{covid_stance_bert_performance}
    \begin{tabular}{c c c}
    \toprule
    Target & Acc & F1\\
    \midrule
    Anthony S. Fauci, M.D. & 0.7714 & 0.7557\\
    Stay at Home Orders & 0.8652 & 0.8340\\
    Wearing a Face Mask & 0.8257 & 0.8180\\
    \bottomrule
\end{tabular}
\end{table}
\subsection{Illustrations of Opinion Diversity}
\label{sec:illustration_opinion_diversity}

Different illustrations of opinion diversity calculation can be found in Table~\ref{opinion_f1_table}. 
The opinion precision measures the proportion of important opinions in the generated summary. The opinion recall measures the degree of salient opinion in the source documents that the generated summary contains.

\begin{table*}[tbh]
\caption{Illustration of different opinion precision and recall scenarios.}
    \label{opinion_f1_table}
    \centering
    {\fontsize{10}{12}\selectfont
    \begin{tabular}{|llll|}
    \hline
    \multicolumn{4}{|l|}{Source documents: \textbf{Opinion A}; \textbf{Opinion B}}                                                                                  \\ \hline
    \multicolumn{1}{|l|}{Model generated summary:}                          & \multicolumn{1}{l|}{Opinion Precision} & \multicolumn{1}{l|}{Opinion Recall} & F1   \\ \hline
    \multicolumn{1}{|l|}{Good precision, weak recall: \textbf{Opinion A}} & \multicolumn{1}{l|}{1.00}      & \multicolumn{1}{l|}{0.50}   & 0.67 \\ \hline
    \multicolumn{1}{|l|}{Good precision, good recall: \textbf{Opinion A}; \textbf{Opinion B}} & \multicolumn{1}{l|}{1.00}      & \multicolumn{1}{l|}{1.00}   & 1.00 \\ \hline
    \multicolumn{1}{|l|}{Weak precision, weak recall: \textbf{Opinion A}; Opinion C} & \multicolumn{1}{l|}{0.50}      & \multicolumn{1}{l|}{0.50}   & 0.50 \\ \hline
    \multicolumn{1}{|l|}{Bad precision, bad recall: Opinion C}   & \multicolumn{1}{l|}{0.00}      & \multicolumn{1}{l|}{0.00}   & 0.00 \\ \hline
\end{tabular}}
\end{table*}

\begin{table*}[t]
\caption{An example of source tweets and generated summaries with different opinions towards the "Stay at Home Orders" topic. In the source tweets, users expressed support, against and neutral stances towards the topic, forming three different opinions. In the generated summaries, ChatGPT covered two opinions, whereas Pegasus \cite{zhang2020pegasus} covered a single opinion. In this scenario, ChatGPT has better opinion diversity.}
\label{example_opinion}
\begin{adjustbox}{max width=\linewidth}
\begin{tabular}{lc}
\toprule
\midrule
\textbf{Source tweets} & \textbf{Stance in source tweets} \\ \midrule
‘Hey, \#billmaher, having people stay home is not “fear,” it’s *public health.* @USER’ \\ 
 ‘@USER It goes against every fibre of my being to stay home but I have to because \\my dads high risk. If you die from covid you can’t support the movement long-term. \\Do what you can from home if you’re worried’
& Support   
\\ \midrule

‘@USER Zero protesters but we all are ordered to stay home, and this is a "free" country.\\ What a disgrace to the veterans that died for "free".’
& Against   
\\ \midrule
‘The UFCW told their members to “please stay home” and expressed their sympathies \\ to members “who lost their workplaces.” And this is in a union that has tens-of-thousands \\of Black members working in essential low-wage jobs in retail and meat packing. Cowards.’                                                                  
& Neutral   \\ 
\midrule
\midrule
\textbf{Generated summaries} & \textbf{Stance in generated summaries} \\
\midrule
\textbf{ChatGPT}\\
‘People are upset about being told to stay home during the pandemic, \\but it's important for public health.'\\
‘Some are frustrated that they can't attend church or receive sacraments, \\while others are protesting and attending funerals.' & Support \\ 
‘There are concerns about the spread of COVID-19 and the impact on vulnerable individuals.'\\
‘Some are critical of those who are not taking' & Against \\
\midrule
\textbf{Pegasus}\\
‘I stay home to protect family, friends, strangers .’\\
 ‘...and this is in a union that has tens-of-thousands of black members working \\in essential low-wage jobs in retail and meat packing.’ \\
‘...and this is in a state that has much harsher lasting restrictions depending on \\ the needs of each state...look it up during a’ & Neutral\\
\midrule
\midrule
\end{tabular}
\end{adjustbox}
\end{table*}

\subsection{Stance Examples and Opinion Diversity Calculations}
\label{sec:example_opinion_diversity_calculation}
Table~\ref{example_opinion} provided an example of the stance expressed in the source tweets and the generated summaries using a cluster of tweets under the topic of "Stay at Home Orders". In this example, the source tweets presented three different stances towards the topic. 
For the generated summaries, Pegasus \cite{zhang2020pegasus} presented only a neutral stance towards the topic, while ChatGPT successfully captured both support and against stances. 
We provide how opinion diversity can be measured using the provided example and the calculation detail can be found in Table~\ref{example_illustration}.
Applying F1 to these sets of stances yields the results shown in Table~\ref{example_opinion}; from this, we can see that ChatGPT in this example has a higher opinion diversity score.

\begin{table}[tbh]
\caption{Illustration of opinion diversity calculation using examples provided in Table~\ref{example_opinion}}
\label{example_illustration}
\begin{tabular}{l}
    \toprule
    \textbf{Source tweets}: Stance-A, Stance-B, Stance-C\\
    \midrule
    \textbf{System summary}:\\
    ChatGPT: Stance-A, Stance-B \\
    Pegasus: Stance-C \\
    \midrule
    \textbf{Opinion diversity}:\\
    ChatGPT: Precision = 2/2; Recall = 2/3; F1 =  0.8\\
    Pegasus: Precision = 1/1; Recall = 1/3; F1 =  0.5\\
    \bottomrule
\end{tabular}
\end{table}


\subsection{Additional Results}
\label{sec:additional_results}

\begin{table*}[tbh]
  \caption{Results of opinion precision and opinion recall for various models under different discussed topics.}
  \label{opinion_precision_recall}
    \centering
    {\fontsize{10}{12}\selectfont
    \begin{tabular}{c c c c c c c}
    \toprule
     Events & 
    \multicolumn {2}{c}{CDC} &
    \multicolumn {2}{c}{Stay at Home Orders} &
    \multicolumn {2}{c}{Wearing a Face Mask} \\
    \midrule
    Models & Precision & Recall & Precision & Recall & Precision & Recall\\
    \midrule
        BART	&	\textbf{0.8462}	&	0.7286	&	\textbf{1.0000}	&	0.6354	&	\textbf{0.9808}	&	\textbf{0.7308}	\\
        Pegasus	&	0.6111	&	0.5256	&	0.9931	&	0.6424	&	0.4071	&	0.3558	\\
        T5	&	0.7137	&	0.6560	&	0.9826	&	0.6285	&	0.4872	&	0.4615	\\
        ChatGPT	&	0.7863	&	\textbf{0.7521}	&	0.9931	&	\textbf{0.6979}	&	0.8077	&	0.5000	\\
        Copycat	&	0.6303	&	0.5192	&	0.9792	&	0.5868	&	0.8782	&	0.5929	\\
        TextRank	&	0.6303	&	0.5235	&	0.9931	&	0.6319	&	0.2981	&	0.2468	\\
        LexRank	&	0.6517	&	0.5577	&	\textbf{1.0000}	&	0.6389	&	0.3942	&	0.3397	\\
        Hybrid TFIDF	&	0.6496	&	0.5791	&	0.9792	&	0.5833	&	0.4038	&	0.3494	\\
    \bottomrule
\end{tabular}}
\end{table*}

Table~\ref{opinion_precision_recall} contains results of opinion precision and recall for the three discussed topics. Based on the result, we found that BART \cite{lewis2020bart} has the best precision score across the three discussed topics. ChatGPT has relatively strong recall scores. BART is better at capturing important opinions, while ChatGPT is relatively better at capturing salient opinions.

\end{document}